\documentclass[11pt,a4paper]{article}
\usepackage[whole]{bxcjkjatype}
\usepackage[hyperref]{naaclhlt2019}
\usepackage{times}
\usepackage{latexsym}
\usepackage{amsfonts}
\usepackage{amsmath}
\usepackage{graphicx}

\usepackage{url}
\usepackage{subfig}
\usepackage{comment}

\aclfinalcopy 
\def\aclpaperid{943} 

\title{Multi-task Learning for Japanese Predicate Argument Structure Analysis}

\author{Hikaru Omori \\
  Tokyo Metropolitan University \\
  {\tt omori-hikaru@ed.tmu.ac.jp} \\\And
  Mamoru Komachi \\
  Tokyo Metropolitan University \\
  {\tt komachi@tmu.ac.jp} \\}

\date{}

\begin{document}
\maketitle
\begin{abstract}
An {\it event-noun} is a noun that has an argument structure similar to a predicate. 
Recent works, including those considered state-of-the-art, ignore event-nouns or build a single model for solving both Japanese predicate argument structure analysis (PASA) and event-noun argument structure analysis (ENASA).
However, because there are interactions between predicates and event-nouns, it is not sufficient to target only predicates.
To address this problem, we present a multi-task learning method for PASA and ENASA.
Our multi-task models improved the performance of both tasks compared to a single-task model by sharing knowledge from each task.
Moreover, in PASA, our models achieved state-of-the-art results in overall F1 scores on the NAIST Text Corpus.
In addition, this is the first work to employ neural networks in ENASA.
\end{abstract}

\section{Introduction}
Japanese predicate argument structure analysis (PASA) examines semantic structures between the predicate and its arguments in a text.
The identification of the argument structure such as ``who did what to whom?" is useful for natural language processing that requires deep analysis of complicated sentences such as machine translation and recognizing textual entailment.
PASA is a task targeted at predicates such as verbs and adjectives. 
However, there are also many nouns that have event-related arguments in a sentence. 
We call these nouns that refer to events {\it event-nouns}, for example, a verbal noun (\textit{sahen} nouns) such as {\it houkoku} ``report" or a deverbal noun (nominalized forms of verbs) such as {\it sukui} ``rescue."

\begin{figure}[t]
\centering
\subfloat[][He reports the result to his boss.]{\includegraphics[width=48mm]{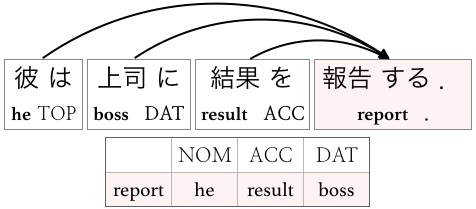}\label{e1}}
\\
\subfloat[][His progress report was too short; hence, he got scolded by his boss.]{\includegraphics[width=70mm]{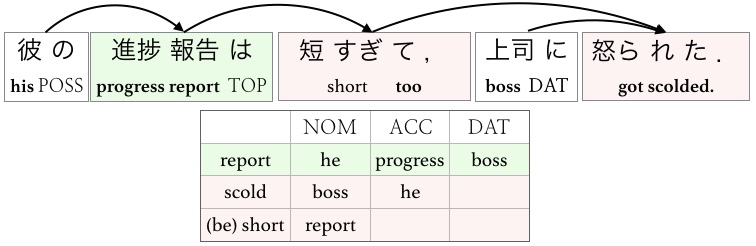}\label{e2}}
\\
\subfloat[][He sent a brief note to his boss.]{\includegraphics[width=53mm]{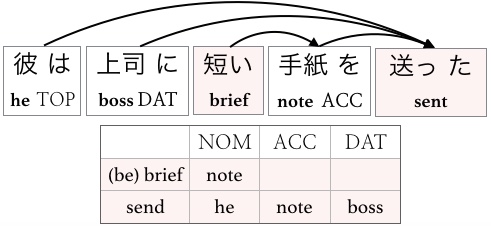}\label{e3}}
\caption[]{Examples of PASA and ENASA. The edges denote dependency paths.}
\label{example}
\end{figure}

Figure \ref{example} shows examples of PASA and event-noun argument structure analysis (ENASA). 
In the NAIST Text Corpus \cite{NTC_English}, both predicates and event-nouns have one of three core case roles, {\it nominative} (NOM), {\it accusative} (ACC), and {\it dative} (DAT) as an argument.
According to \newcite{NTC_English}, predicates have almost no argument in the
same {\it bunsetsu}\footnote{Functional chunk in Japanese. It consists of one
or more content words (noun, verb, adjective, etc.) followed by zero or more
function words (postposition, auxiliary verb, etc.). A verb phrase in Japanese thus cannot bear noun arguments in the same bunsetsu.} phrase. However, in the case of event-nouns, approximately half of the accusative and dative arguments appear in the same bunsetsu phrase.
Accordingly, although PASA and ENASA are semantically highly related, they are syntactically different tasks.
However, most previous studies focused on predicates only; hence, there are few studies that focus on event-nouns \cite{Komachi-PACLING,Taira}. 
To identify the semantic units of a sentence and to correctly understand syntactic relations, it is not sufficient to target only PASA.

Thus, we propose a multi-task learning model that effectively leverages ENASA and improves PASA.  
Our proposed model is based on an end-to-end multilayer bi-directional recurrent neural network (RNN) used in recent works, and the model has networks that distinguish task-independent information and task-specific information.

In summary, the main contributions of this work are the following: 

\begin{enumerate}
\item This is the first attempt to design a multi-task learning framework for PASA and ENASA, and we show that our models improve the performance of both tasks. 
\item Although our model is a simple model that does not consider the interactions between multiple predicates, it achieves a state-of-the-art result on the NAIST Text Corpus (NTC) in PASA by combining syntactic information as one of the features.
\item For ENASA, this is the first work to employ neural networks to effectively incorporate PASA.
\end{enumerate}

\section{Related Work}

\subsection{Japanese PASA and ENASA Approaches}
Many machine learning-based methods have been studied in Japanese PASA.
Traditional models take pointwise approaches that construct independent models for each core case role (NOM, ACC, DAT).
\newcite{Taira} proposed a supervised model that learns features of each case using decision lists and support vector machines. 
\newcite{Imamura} proposed a model that combines a maximum entropy model with a language model trained from large-scale newspaper articles.
\newcite {Hayashibe} designed three models exploiting argument position and type and determined the maximum likelihood output using pairwise comparison.

However, the joint approach that optimizes the scores of all predicate-argument pairs in a sentence simultaneously showed better results than the pointwise approach.
\newcite{Yoshikawa} proposed a model that considers dependency between multiple predicate-argument relations using Markov logic networks.
\newcite{Ouchi2015} jointly optimized the combinations among multiple predicates and arguments in a sentence using a bipartite graph. 

Except for \cite{Taira}, these studies focused on the analysis of predicates while there are few studies that focus on event-nouns.
\newcite{Komachi-PACLING} decomposed ENASA into two tasks: {\it event-hood}
determination and argument identification; they proposed a supervised method using lexico-syntactic patterns. 
Event-hood determination is the most important characteristic that semantically differentiates ENASA from PASA. 
It is a task to determine whether a noun refers to an event (e.g., \textit{houkoku} can refer to either ``to report'' or the outcome of reporting action, ``a report'').
Since the previous ENASA models adopted the pointwise approach with a single model, they did not explore the effective features in each task.
In contrast, our models simultaneously optimize three core case roles. Moreover, the proposed models allow us to distinguish between task-shared and task-specific features using multi-task learning.

\subsection{PASA using neural networks}
Some neural models have achieved higher performance than traditional machine learning models in Japanese PASA.
\newcite{Shibata} replaced \newcite{Ouchi2015}'s scoring function with feed forward neural networks.
\newcite{Matsubayashi} represented a dependency path between a predicate and its argument with path embeddings and showed that even the local model without multiple predicates can outperform a global model. 

Moreover, some end-to-end models have been proposed in Japanese PASA.
\newcite{Ouchi2017} proposed an end-to-end model based on the model using eight-layer bi-directional long short-term memory (LSTM) proposed by \newcite{Zhou-Xu} and considered the interaction of multiple predicates simultaneously using a Grid RNN.
\newcite{Matsubayashi2018} combined self-attention with \newcite{Ouchi2017}'s model to directly capture interaction among multiple predicate-arguments.
In particular, the model improved the performance of arguments that have no syntactic dependency with predicates and achieved a state-of-the-art result on Japanese PASA.

\subsection{Semantic Role Labeling}
Semantic role labeling (SRL) is a similar task to Japanese PASA.
Recently, several end-to-end models using neural networks showed high performance in English SRL \cite{Zhou-Xu,He,Tan}.
\newcite{SRL-SOTA} proposed a multi-task learning model that jointly learned dependency parsing, part-of-speech tagging, predicate detection, and SRL based on multi-head self-attention.
\newcite{Ouchi2018} proposed a span-based SRL model using bi-directional LSTMs and achieved state-of-the-art results. 
The authors scored all possible spans for each label and selected correct spans satisfying constraints when decoding.
In terms of the event-noun research, \newcite{Gerber} used pointwise mutual information (PMI) as a feature for 10 event-nouns with high frequency and identified semantic roles using a logistic regression model.

There were several LSTM models that also achieved high accuracy gains in Chinese SRL \cite{Wang-CSRL,Roth-CSRL,Sha-CSRL,Marcheggiani-CSRL,Qian-CSRL}.
For event-nouns, \newcite{Li_Chinese} showed that combining effective features in verbal SRL with nominal SRL can improve results.
Although the authors did not demonstrate that verbal SRL also improves performance in combination with nominal SRL, we show that our model improves performance in both PASA and ENASA.

\section{Japanese PASA and ENASA}
\begin{figure*}[h]
\centering
\includegraphics[width=158mm]{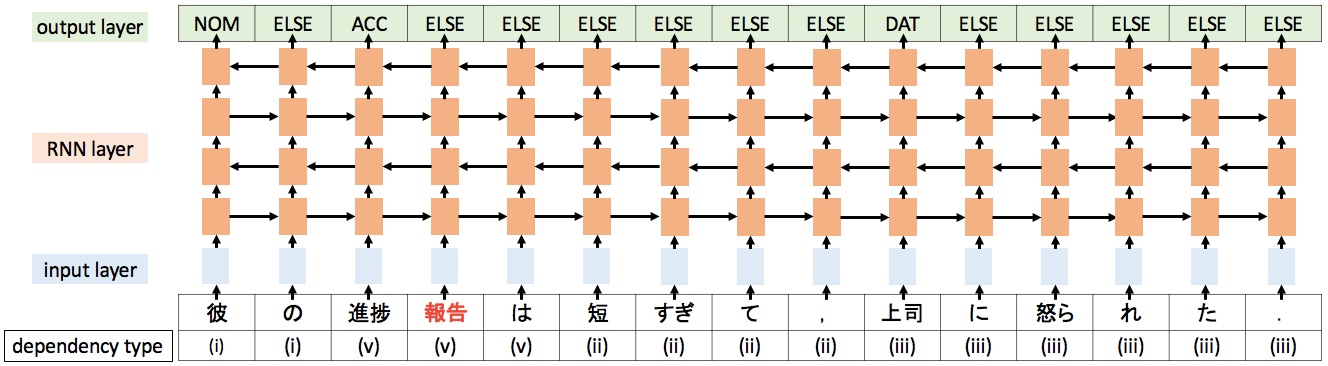}
\caption{End-to-end single model.}
\label{model}
\end{figure*}

\subsection{Task Description}
Japanese predicate (event-noun) argument structure analysis is a task to extract arguments for certain predicates (event-nouns) and assign three case labels, NOM, ACC and DAT \cite{NTC_English}.  
Arguments are divided into four categories \cite{Taira} according to the positions with their predicates (event-nouns).

\begin{description}
\item[Dep]Arguments depend on their predicate (event-noun), or a predicate (event-noun) depends on its arguments.
\item[Zero]Arguments and their predicate (event-noun) are in the same sentence, but the arguments are omitted by zero anaphora. Therefore, they have no direct dependency.
\item[Inter-zero]Zero anaphoric arguments and their predicate (event-noun) are not in the same sentence.
\item[Bunsetsu]Arguments and their event-noun are in the same bunsetsu.

\end{description}
A sentence $w=w_1,w_2,\cdots,w_T$ and a predicate (event-noun) $p=p_1,p_2,\cdots,p_q$ are given as input.
\newcite{Iida2006}, \newcite{Imamura}, and \newcite{Sasano} also analyze Inter-zero, which is a difficult task because the whole document must be searched. 
Following existing research \cite{Ouchi2015,Ouchi2017,Matsubayashi,Matsubayashi2018,Taira}, we only focus on three categories where arguments and their predicate (event-noun) are in the same sentence.
In addition, we exclude the Bunsetsu category from the PASA evaluation following \newcite{Ouchi2017} and \newcite{Matsubayashi2018}.

\subsection{End-to-end Single Model}
Our single model is based on an end-to-end approach \cite{Zhou-Xu,Ouchi2017,Matsubayashi2018}. Additionally, we add new features.
Figure \ref{model} shows the network architecture of our base model.

\subsubsection{Input Layer}
Each word $w_t\in[w_1,\cdots,w_T]$ is converted to a feature representation ${\bf x}_t\in[{\bf x}_1,\cdots,{\bf x}_T]$ at the input layer. 
We use six types of features. 
The feature representation ${\bf x}_t$ is defined as follows: 
\begin{eqnarray}
  {\bf x}_t &=& {\bf x}_t^{\mathrm {as}} \oplus {\bf x}_t^{\mathrm {posi}} \oplus {\bf x}_t^{\mathrm {dep}} \oplus {\bf x}_t^{\mathrm {type}} \oplus {\bf x}_p^{\mathrm {task}}
\end{eqnarray}
where ($\oplus$) indicates concatenation of vectors.
\paragraph{Argument Structure}
Predicate (event-noun) $w_p$ and argument candidates $w_t$ are converted to the vectors ${\bf x}_t^{\mathrm {as}}\in\mathbb{R}^{2d_{\mathrm {w}}}$ by the word embedding matrix.

\paragraph{Position}
This is a feature that represents the positional relation between $w_p$ and $w_t$. 
The feature is calculated by subtracting the word index of argument candidates from the word index of predicates (event-nouns).
We use two types of units to represent relative position: word unit $p^{\mathrm {word}}_t$ and bunsetsu unit $p^{\mathrm {bunsetsu}}_t$, which are converted to the word positional vector ${\bf p}_t^{\mathrm {word}}\in\mathbb{R}^{d_{\mathrm {p}}}$ and the bunsetsu positional vector ${\bf p}_t^{\mathrm {bunsetsu}}\in\mathbb{R}^{d_{\mathrm {p}}}$, respectively, by the word and bunsetsu positional embedding matrices. 
We concatenate these two vectors and obtain the positional vectors ${\bf x}_t^{\mathrm {posi}}\in\mathbb{R}^{2d_{\mathrm {p}}}$.

\paragraph{Dependency}
This is a feature that represents the dependency relation between $w_p$ and $w_t$.
We set five types of dependency relations:
\begin{enumerate}
\renewcommand{\labelenumi}{\roman{enumi}).}
\item Argument candidates depend on the predicate (event-noun). 
\item The predicate (event-noun) depends on the argument candidates.
\item No dependency relations between the predicate (event-noun) and argument candidates.
\item The predicate and candidate arguments are in the same bunsetsu.
\item The event-noun and candidate arguments are in the same bunsetsu.
\end{enumerate}
The dependency relation $d_t$ is converted to the dependency vector ${\bf x}_t^{\mathrm {dep}}\in\mathbb{R}^{d_{\mathrm {d}}}$ by the dependency relation embedding matrix. 
The dependency type in Figure \ref{model} shows how to make dependency features in Figure \ref{e2} as an example.
We define the dependency type from the syntactic information annotated in the NTC.

In previous work, dependency features are used differently from our study.
\newcite{Imamura} used a binary feature that represents whether or not there is a dependency relation between the predicate and its arguments.
We employ more fine-grained relation types to adapt to event-nouns.
\newcite{Matsubayashi} represented the interactions between a predicate and its arguments using path embedding.
In contrast, we define different types for a predicate and event-noun to distinguish event-nouns from predicates and learn embeddings to find the associated latent structures.

\paragraph{Event-hood Type}
This is a binary feature to flag all predicates (event-nouns) in a sentence inspired by \newcite{Matsubayashi2018}.
The purpose of this feature is to prevent predicates from becoming arguments and to help some event-nouns become arguments.
The event-hood type vector ${\bf x}_t^{\mathrm {type}}\in\mathbb{R}^{2}$ of a candidate indicates [0,1] if the candidate is a predicate, [1,0] if the candidate is an event-noun, and [0,0] otherwise. 
The predicate and event-noun are annotated in the NTC.

\paragraph{Task Label}
This is a binary feature vector ${\bf x}_p^{\mathrm {task}}\in\mathbb{R}^{1}$ that indicates 1 if the task is predicate argument structure analysis; otherwise, 0.

\subsubsection{RNN Layer}
We use the gated recurrent unit (GRU) \cite{GRU} for RNN. The RNN layers are made up of $L$ layers of stacked bi-directional GRU.  
Additionally, we apply the residual connections \cite{Residual} following \newcite{Ouchi2017,Matsubayashi2018}.
At each time step $t$, the hidden state ${\bf h}^l_t\in\mathbb{R}^{d_{\mathrm {h}}}$ in the $l\in[1,\cdots,L]$-th layer is calculated as follows: 
\begin{eqnarray}
  {\bf h}^l_t &=& 
  \begin{cases}
    g^l({\bf h}^{l-1}_t, {\bf h}^l_{t-1}) & (l=\rm{odd}) \\
    g^l({\bf h}^{l-1}_t, {\bf h}^l_{t+1}) & (l=\rm{even})
  \end{cases}
\end{eqnarray}
where $g^l(\cdot)$ denotes the $l$-th layer GRU function. In addition, ${\bf h}^0_t={\bf x}_t$. 

\begin{figure*}[t]
\centering
\subfloat[][Single]{\includegraphics[width=45mm]{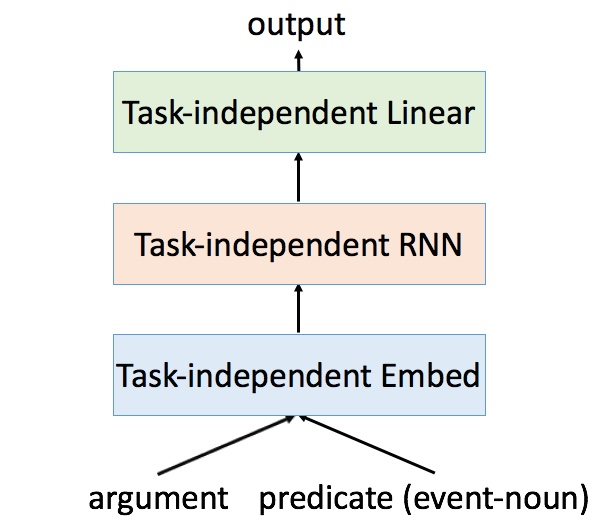}\label{single}} \quad
\subfloat[][Multi-input]{\includegraphics[width=57.5mm]{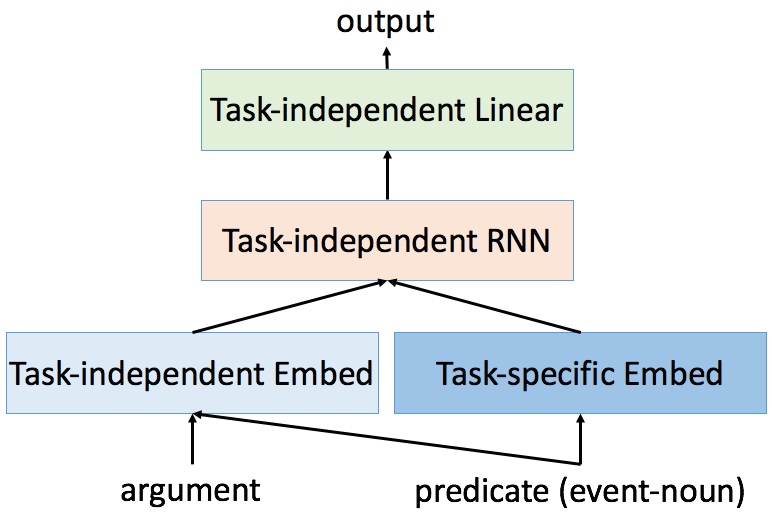}\label{emb}} \quad
\subfloat[][Multi-RNN]{\includegraphics[width=45mm]{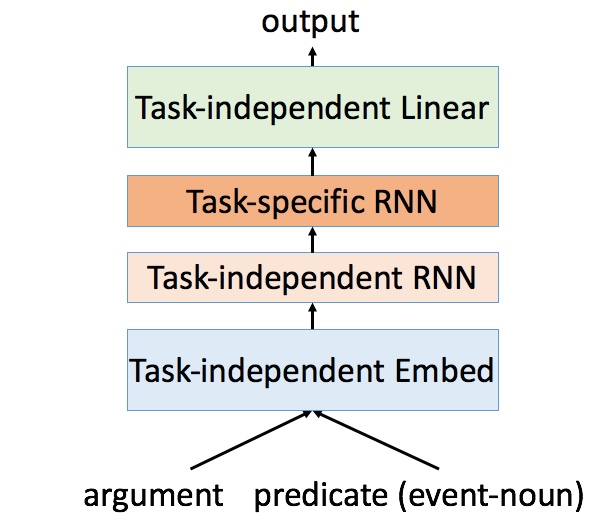}\label{rnn}} \quad
\subfloat[][Multi-output]{\includegraphics[width=57.5mm]{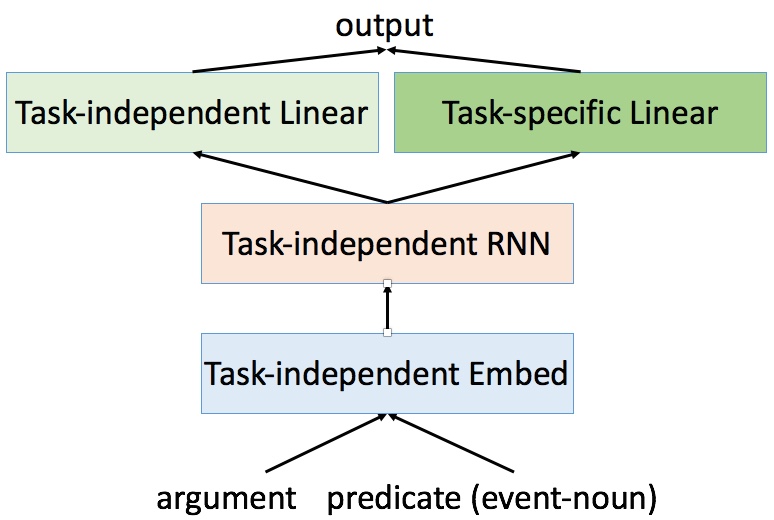}\label{linear}} \quad
\subfloat[][Multi-ALL]{\includegraphics[width=57.5mm]{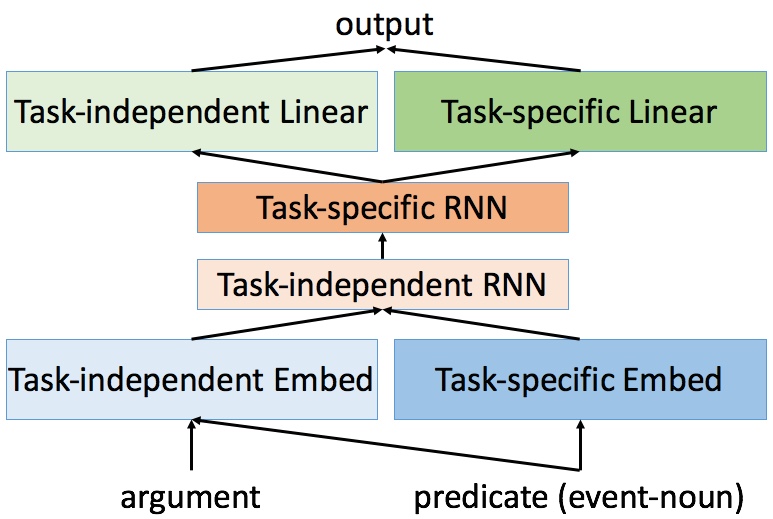}\label{erl}} \quad
\caption[]{Proposed models: \subref{single} Single, \subref{emb} Multi-input, \subref{rnn} Multi-RNN, \subref{linear} Multi-output, \subref{erl} Multi-ALL.}
\label{multi}
\end{figure*}

\subsubsection{Output Layer}
In the output layer, we input each hidden state ${\bf h}^L_t$. Then, we obtain the output vector ${\bf o}_t$ using the softmax function:
\begin{eqnarray}
  {\bf o}_t & = & {\rm softmax}({\bf W}_\mathrm{o} {\bf h}^L_t+{\bf b}_\mathrm{o}) 
\end{eqnarray}
where ${\bf W}_\mathrm{o}\in\mathbb{R}^{4\times d_{\mathrm {h}}}$ is the parameter matrix, and ${\bf b}_\mathrm{o}\in\mathbb{R}^{4}$ is the bias term.
The output vector represents the probability for each argument candidate over four labels, [NOM, ACC, DAT, ELSE].
ELSE denotes that the candidate argument does not have a case label.
In testing, the maximum probability label is selected as the output label.
We train the model using the cross-entropy loss function.

\section{Multi-task Model}

Multi-task learning has been successfully applied to various natural language processing tasks \cite{MTL-Collobert,MTL-Sogaard,MTL-Luong,MTL-Hashimoto,MTL-Liu,M-M-task,MTL-ORL,SRL-SOTA}.
One of the advantages of multi-task learning is that it learns better representation, which is robust against task-dependent noise by increasing training data. 
In this paper, we introduce multi-task learning to PASA and ENASA for the first time.
We propose three methods to extend the end-to-end single model to the multi-task learning model in the input layer, RNN layer, and output layer. Our final model combines all three methods.

\subsection{Multi Input Layer}
Even if the surface form is the same, the contexts are different for predicates and event-nouns.
For example, the event-noun {\it houkoku} ``report" in Figure \ref{e2} has an argument in the same bunsetsu unlike predicates.
Moreover, the event-noun also has a nominative argument role for the predicate {\it mijikai} ``short".
Therefore, given this, we prepare a task-specific word embedding matrix that addresses the task-specific distribution of words.
The predicate is converted to PASA-specific vectors ${\bf x}_t^{\mathrm{p}}\in\mathbb{R}^{d'_{\mathrm {w}}}$ by the PASA-specific predicate embedding matrix.
Similarly, the event-noun is converted to ENASA-specific vectors ${\bf x}_t^{\mathrm{n}}\in\mathbb{R}^{d'_{\mathrm {w}}}$ by the ENASA-specific event-noun embedding matrix.
These matrices are randomly initialized and can be learned during training.

The feature vector ${\overline {\bf x}_t}$ is defined as follows:
\begin{eqnarray}
  {\overline {\bf x}_t} &=&
  \begin{cases}
  {\bf x}_t \oplus {\bf x}_t^{\mathrm{p}} & (\rm{PASA}) \\
  {\bf x}_t \oplus {\bf x}_t^{\mathrm{n}} & (\rm{ENASA})
  \end{cases}
\end{eqnarray}

\subsection{Multi RNN Layer}
Previous work \cite{MTL-Sogaard,MTL-Hashimoto} proposed hierarchical multi-task learning models that exploited features obtained from easy tasks for difficult tasks.
These studies showed that performance improves when low-layer RNN representations are trained in easy tasks and high-layer RNN are leveraged for difficult tasks.
Therefore, we construct a network that hierarchically overlaps a task-specific RNN on a task-independent RNN.
Lower RNN layers learn task-independent knowledge representations. Then, the task-specific RNN adjusts the representations for each task. 
At each time step $t$, the hidden state ${\bf m}^{l'}_t\in\mathbb{R}^{d'_{\mathrm {h}}}$ in the $l'\in[1,\cdots,L']$-th layer is calculated as follows: 
\begin{eqnarray}
  {\bf m}^{l'}_t &=& 
  \begin{cases}
    g^{l'}({\bf m}^{l'-1}_t, {\bf m}^{l'}_{t-1}) & ({l'}=\rm{odd}) \\
    g^{l'}({\bf m}^{l'-1}_t, {\bf m}^{l'}_{t+1}) & ({l'}=\rm{even})
  \end{cases}
\end{eqnarray}
\begin{eqnarray}
  g^{l'}(\cdot) &=&
  \begin{cases}
    g_{\mathrm{p}}^{l'}(\cdot) & (\rm{PASA})  \\
    g_{\mathrm{n}}^{l'}(\cdot) & (\rm{ENASA})
  \end{cases}
\end{eqnarray}
where $g^{l'}(\cdot)$, $g_{\mathrm{p}}^{l'}(\cdot)$, and $g_{\mathrm{n}}^{l'}(\cdot)$ denote the $l'$-th layer GRU functions. In addition, ${\bf m}^0_t={\bf h}_t^L$.

\subsection{Multi Output Layer}
The position of arguments is different with respect to predicates and event-nouns.
For example, predicates seldom have arguments in the same bunsetsu. In contrast, event-nouns often have arguments in the same bunsetsu, compound nouns, for example.
Therefore, it is intuitive and natural to divide the output layer into task-independent and task-specific layers.
The task-specific output vectors are calculated as follows: 
\begin{eqnarray}
  {\bf o}_t^{\mathrm {p}} & = & {\bf W}_{\mathrm {o}}^{\mathrm {p}} {\bf h}_t+{\bf b}_{\mathrm {o}}^{\mathrm {p}} \\
  {\bf o}_t^{\mathrm {n}} & = & {\bf W}_{\mathrm {o}}^{\mathrm {n}} {\bf h}_t+{\bf b}_{\mathrm {o}}^{\mathrm {n}} \\
  {\bf g}_t & = & \sigma({\bf W}_{\mathrm {g}} {\bf h}_t+{\bf b}_{\mathrm {g}})
\end{eqnarray} 
where ${\bf W}_{\mathrm {o}}^{\mathrm {p}},{\bf W}_{\mathrm {o}}^{\mathrm {n}},{\bf W}_{\mathrm {g}}\in\mathbb{R}^{4\times d_{\mathrm {h}}}$ are the parameter matrices, and ${\bf b}_{\mathrm {o}}^{\mathrm {p}}, {\bf b}_{\mathrm {o}}^{\mathrm {n}}, {\bf b}_{\mathrm {g}}\in\mathbb{R}^{4}$ are the bias terms. 
${\bf h}_t$ is the hidden state of the last layer.
We combine task-specific output vectors ${\bf o}_t^{\mathrm {p}},{\bf o}_t^{\mathrm {n}}$ with task-independent output vector ${\bf o}_t$ by the gate ${\bf g}_t$.
\begin{eqnarray}
  {\bf c}_t &=&
  \begin{cases}
    {\bf g}_t \odot {\bf o}_t + (1-{\bf g}_t) \odot {\bf o}_t^{\mathrm {p}} & (\rm{PASA})  \\
    {\bf g}_t \odot {\bf o}_t + (1-{\bf g}_t) \odot {\bf o}_t^{\mathrm {n}} & (\rm{ENASA})
  \end{cases} \\
  {\overline {\bf o}_t} &=& {\rm softmax}({\bf c}_t)
\end{eqnarray}
where ($\odot$) denotes the element-wise product.
The output vector ${\overline {\bf o}_t}$ represents the probability of [NOM, ACC, DAT, ELSE]. 

\section{Experiments}
\subsection{Dataset and Setting}
We use NTC 1.5 for our experiments.
We divide the dataset into training, development, and test sets in the same way as \newcite{Taira}.
We use morphological and syntactic information, such as the word boundaries, the bunsetsu boundaries and the dependency relations provided in the NTC.

For the development and test sets, if there are two or more arguments annotated with the same case label in a sentence, we set an argument that only has a dependency relation with a predicate as a correct answer and assign the ELSE label to other arguments. 
If there is no dependency relation, we set an argument with the shortest distance $|w_p-w_t|$ as a correct answer.
If the distance is equal, an argument on the left side of a predicate is considered a correct answer.

In NTC 1.5, if there is a predicate phrase, such as ``verbal noun + {\it suru}," {\it suru} is annotated as a predicate word.  
We consider the verbal noun as the predicate word at the preprocessing step to match the surface of a predicate with that of an event-noun. 
Take the predicate {\it houkoku-suru} ``to report" and an event-noun {\it houkoku} ``report" as an example.
Although $w_p$ before preprocessing are {\it suru} and {\it houkoku},  $w_p$ are unified to {\it houkoku} after preprocessing.

\subsection{Hyperparameters}
\begin{table}[t]
\centering
\scalebox{0.82}{
\begin{tabular}{| l | r |} \hline
the dimension of word embeddings $d_{\mathrm {w}}$ & 300 \\
the dimension of position embeddings $d_{\mathrm {p}}$ & 16 \\
the dimension of dependency embeddings $d_{\mathrm {d}}$ & 16 \\
the dimension of hidden states $d_{\mathrm {h}}$ & 300 \\
the number of GRU layers $L$ & 4 \\
the dimension of task-specific word embeddings $d'_{\mathrm {w}}$ & 16 \\
the dimension of task-specific hidden states $d'_{\mathrm {h}}$ & 300 \\
the number of task-specific GRU layers $L'$ & 2 \\
dropout rate & 0.4 \\
batch size & 8 \\
gradient clipping & 4 \\ \hline
\end{tabular}
}
\caption{Hyperparameters.\label{hyper}}
\end{table} 

We use pre-trained embeddings\footnote{http://www.asahi.com/shimbun/medialab/word\_embedding} for the initial values of the word embedding matrix.
The initial values of the other embedding matrices are sampled according to a uniform distribution of [-0.25,0.25].
We convert words appearing more than once in the training set into word vectors and the remaining words into the unknown word vector.  
We adopt AdaDelta \mbox($\epsilon = 10^{-6}$，$\rho = 0.95$) as the optimization method. 
We set the number of epochs to 20 and evaluate the model with the highest F1 scores on the development set.
Table \ref{hyper} shows the hyperparameters.

\subsection{Results}

\begin{table*}[h]
\centering
\scalebox{0.87}[0.87]{
\begin{tabular}{l || c c || c c c c || c c c c} \hline
 & & & \multicolumn{4}{c||}{{\sl Dep}} & \multicolumn{4}{c}{{\sl Zero}} \\   
Method                       &  ALL  &      SD     &  ALL  &  NOM  &  ACC  &  DAT  &  ALL  &  NOM  &  ACC  &  DAT \\
\hline
Ouchi+ 17                    & 81.42       &             & 88.17       & 88.75       & 93.68       & 64.38       & 47.12       & 50.65       & 32.35       & 7.52 \\
M\&I 17                      & 83.50       & $\pm{0.17}$ & 89.89       & 91.19       & 95.18       & 61.90       & 51.79       & 54.69       & 41.8        & 17 \\
M\&I 18                      & 83.94       & $\pm{0.12}$ & 90.26       & 90.88       & 94.99       & 67.57       & {\bf 55.55} & {\bf 57.99} & {\bf 48.9}  & {\bf 23} \\
\hline
Single                       & 83.62       & $\pm{0.17}$ & 90.09       & 90.45       & 94.84       & 69.77       & 51.87       & 54.73       & 43.48       & 11.40  \\
Multi-input                  & 83.88       & $\pm{0.11}$ & 90.27       & 90.65       & 95.12       & 69.86       & 53.01       & 55.82       & 44.68       & 10.77  \\
Multi-RNN                    & 83.91       & $\pm{0.23}$ & 90.17       & 90.58       & 95.07       & 67.94       & 53.31       & 55.85       & 45.71       & 9.97   \\
Multi-output              & 83.77       & $\pm{0.20}$ & 90.13       & 90.68       & 94.89       & 68.16       & 53.93       & 56.73       & 43.79       & 9.45  \\
Multi-ALL                    & 83.82       & $\pm{0.10}$ & 90.15       & 90.68       & 95.06       & 67.56       & 53.50       & 56.37       & 45.36       & 8.70  \\ 
\hline
Multi-RNN+${\tt DEP}$     & 84.55       & $\pm{0.11}$ & 90.69       & 91.28       & 95.25       & 70.07       & 51.56       & 54.29       & 42.67       & 1.85   \\
Multi-output+${\tt DEP}$  & 84.73       & $\pm{0.11}$ & 90.82       & {\bf 91.46} & 95.29       & 70.69       & 52.29       & 55.14       & 42.15       & 1.81  \\
Multi-ALL+${\tt DEP}$     & {\bf 84.75} & $\pm{0.16}$ & {\bf 90.88} & 91.40       & {\bf 95.37} & {\bf 71.02} & 52.35       & 55.10       & 42.54       & 2.32  \\
\hline
\hline
M\&I 17 (ens. of 5)                     & 84.07       &  & 90.24       & 91.59       & 95.29       & 62.61       & 53.66       & 56.47       & 44.7        & 16      \\
M\&I 18 (ens. of 10)                    & 85.34       &  & 91.26       & 91.84       & 95.57       & 70.8        & {\bf 58.07} & {\bf 60.21} & {\bf 52.5}  & {\bf 26} \\
Multi-RNN+${\tt DEP}$ (ens. of 5)    & 85.85       &  & 91.61       & 92.11       & {\bf 95.87} & 72.63       & 53.41       & 55.96       & 46.10       & 0        \\  
Multi-output+${\tt DEP}$ (ens. of 5) & 85.83       &  & 91.52       & 92.12       & 95.69       & 72.72       & 54.35       & 57.02       & 45.95       & 0        \\  
Multi-ALL+${\tt DEP}$ (ens. of 5)    & {\bf 86.01} &  & {\bf 91.63} & {\bf 92.15} & 95.80       & {\bf 72.95} & 54.99       & 57.84       & 45.20       & 0        \\  
\hline
\end{tabular}
}
\caption{F1 scores on the PASA test set. Single is a base model without multi-task learning.}
\label{result1}
\end{table*}

\begin{table*}[h]
\centering
\scalebox{0.68}[0.72]{
\begin{tabular}{l || c c || c c c c || c c c c || c c c c} \hline
 & & & \multicolumn{4}{c||}{{\sl Dep}} & \multicolumn{4}{c||}{{\sl Zero}} & \multicolumn{4}{c}{{\sl Bunsetsu}} \\   
Method                       &  ALL        &      SD     &  ALL        &  NOM        &  ACC        &  DAT        &  ALL        &  NOM        &  ACC        &  DAT        & ALL         &  NOM        &  ACC        &  DAT \\
\hline
Taira+ 08 on NTC 1.4     &             &             &             & 68.01       & 62.46       & 56.05       &             & 36.19       & 20.46       & 6.62        &             & 78.93       & 77.96       & 58.13 \\
\hline
\hline
Single                       & 66.21       & $\pm{0.15}$ & 74.64       & 76.06       & 74.54       & 51.28       & 46.05       & 49.67       & 33.36       & 13.63       & 78.24       & 76.67       & 81.75       & 48.55 \\ 
Multi-input                  & 67.89       & $\pm{0.42}$ & 75.62       & 76.63       & 75.78       & {\bf 57.17} & 49.07       & 52.81       & 36.95       & 19.39       & {\bf 79.35} & 77.31       & {\bf 83.31} & 51.03 \\
Multi-RNN                    & 67.96       & $\pm{0.44}$ & 75.86       & 76.90       & 76.33       & 54.46       & 48.67       & 52.18       & {\bf 38.47} & 18.89       & 79.08       & 77.24       & 82.89       & 50.93 \\
Multi-output                 & 67.96       & $\pm{0.17}$ & 76.25       & {\bf 77.18} & {\bf 76.90} & 54.97       & 48.74       & 52.48       & 36.09       & {\bf 19.64} & 79.02       & 77.00       & 83.04       & 50.60 \\ 
Multi-ALL                    & {\bf 68.00} & $\pm{0.41}$ & 75.90       & 77.16       & 76.05       & 53.00       & {\bf 49.66} & {\bf 53.37} & 37.64       & 14.46       & 79.05       & 77.32       & 82.61       & {\bf 51.83} \\ 
\hline
Multi-ALL+${\tt DEP}$     & 67.68       & $\pm{0.39}$ & {\bf 75.95} & {\bf 77.18} & 76.11       & 55.26       & 47.57       & 51.21       & 35.14       & 15.65       & 79.06       & {\bf 77.44} & 82.66       & 51.10 \\
\hline
\hline
Multi-ALL (ens. of 5)           & {\bf 71.14} &          & {\bf 78.63} & {\bf 79.66} & {\bf 78.83} & 58.29       & {\bf 52.49} & {\bf 56.41} & {\bf 39.02} & 16.42       & {\bf 81.90} & {\bf 80.25} & {\bf 85.21} & {\bf 56.29} \\  
Multi-ALL+${\tt DEP}$ (ens. of 5)  & 69.90 &          & 77.86       & 78.89       & 78.16       & {\bf 58.46} & 49.36       & 53.10       & 36.36       & 17.23       & 81.16       & 79.74       & 84.57       & 52.99 \\
\hline
\end{tabular}
}
\caption{F1 scores on the ENASA test set.}
\label{result2}
\end{table*}

\begin{figure*}[t]
\centering
\subfloat[][predicate: {\it 結成}``organize," NOM: {\it 会長}``president," ACC: {\it 会派} ``faction."]{\includegraphics[width=160mm]{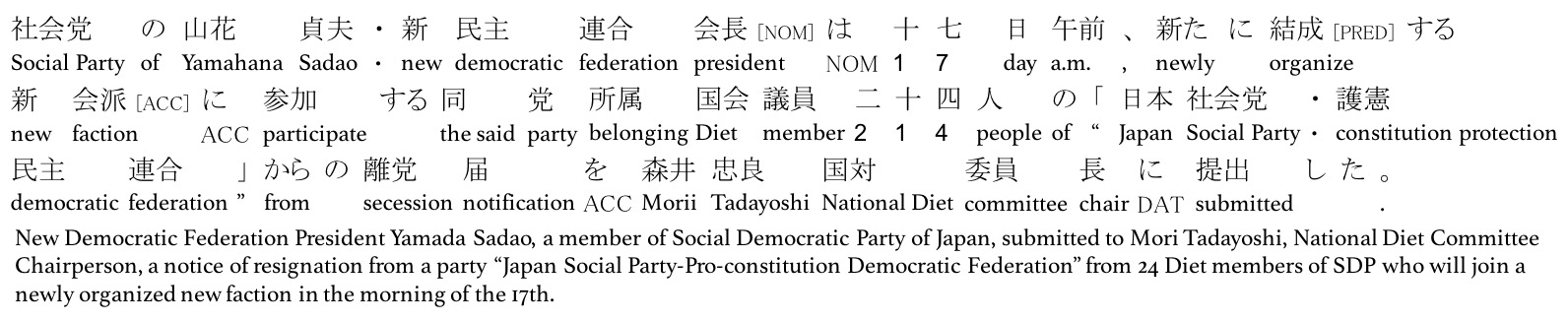}\label{a1}}
\\
\subfloat[][predicate: {\it 左右}``determine," NOM: {\it カギ}``key," ACC: {\it 行方}``whereabouts."]{\includegraphics[width=160mm]{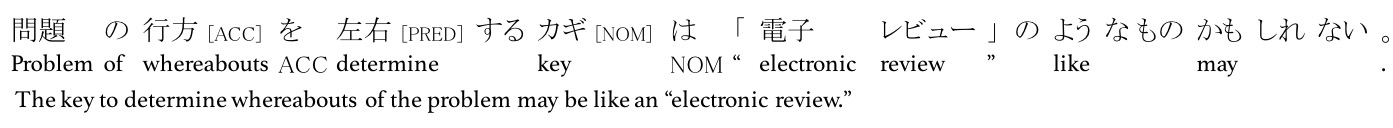}\label{a5}}
\\
\subfloat[][event-noun: {\it 打開}``break," NOM: {\it カギ}``key," ACC: {\it 事態}``situation."]{\includegraphics[width=160mm]{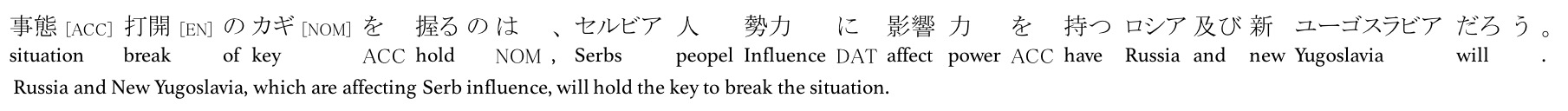}\label{a6}}
\\
\subfloat[][predicate: {\it 回避}``avoid," NOM: {\it トップ}``top," ACC: {\it 責任}``responsibility."]{\includegraphics[width=160mm]{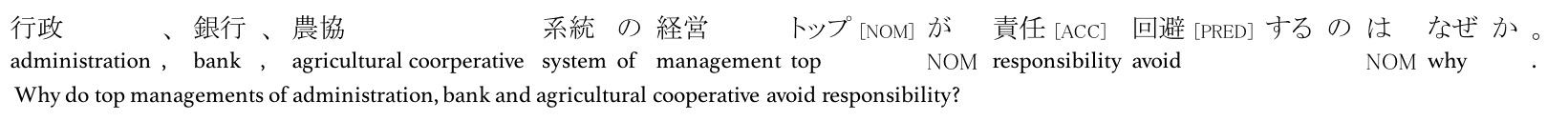}\label{a2}}
\\
\subfloat[][predicate: {\it 押す}``push," ACC: {\it 丸}``circle," DAT: {\it 左下}``lower left."]{\includegraphics[width=160mm]{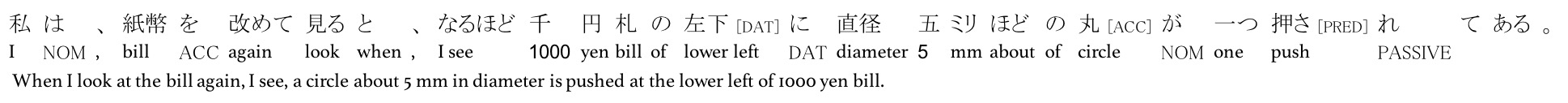}\label{a3}}
\\
\subfloat[][predicate: {\it 果たす}``play," NOM: E, ACC: {\it 役割}``role," DAT: {\it 発病}``pathogenesis."]{\includegraphics[width=160mm]{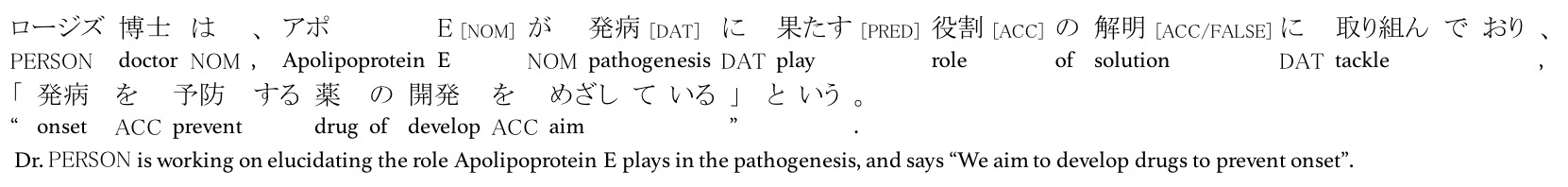}\label{a4}}
\caption[]{Examples of analysis errors on the PASA test set}
\label{analysis}
\end{figure*}

We evaluate each model with the NTC 1.5 test.
The experimental results for the argument structure analysis of predicates and event-nouns are shown in Tables \ref{result1} and \ref{result2}.
\paragraph{Predicate Argument Structure Analysis}
The first set of rows in Table \ref{result1} shows the results of previous models.
Ouchi+ 17 is the model from the Multi-Seq model in \cite{Ouchi2017}.
M\&I 17 is the model in \cite{Matsubayashi}.
M\&I 18 is the model from the MP-POOL-SELFATT model in \cite{Matsubayashi2018}.

The second set of rows in Table \ref{result1} shows the results of the proposed models.
These models do not use the dependency feature.
Compared with the single model, all multi-task learning models improved the overall F1 scores.
Among them, Multi-RNN improved the overall F1 score from the single model by 0.29 points.
In previous work, \newcite{Ouchi2017,Matsubayashi2018} see improvements of 0.27 and 0.55 F1 points in their baseline models by considering multiple predicate-argument interactions.
Therefore, we show that multi-task learning with ENASA achieved comparable effects as these studies in PASA.

The third set of rows shows the results of proposed models using all features including the dependency feature.
Multi-ALL+${\tt DEP}$ achieved the best F1 score among all the models including previous state-of-the-art models.
In particular, the dependency feature was effective for Dep arguments. 
On the other hand, the performance for Zero arguments was poor. 
This result suggests that the dependency feature causes the model to optimize mainly for Dep arguments since Dep arguments are more numerous than Zero arguments.

The fourth set of rows shows the results of ensemble models.
Overall, our proposed model outperformed the previous ensemble model by 0.67 points in the overall F1 score.
Moreover, our models are simple models that independently analyze each predicate in a sentence. Nevertheless, our models achieved higher results than \newcite{Ouchi2017,Matsubayashi2018}.
Although recent works have researched the method whereby multiple predicate-argument interactions are considered simultaneously, how to use syntactic information in the end-to-end model is a subject for future work.

\paragraph{Event-noun Argument Structure Analysis}
The first set of rows in Table \ref{result2} shows the results of a previous model in event-noun argument structure analysis.
Taira+ 08 is the model from \cite{Taira}.
Since its scores are from NTC 1.4, the model cannot be directly compared to our models.
Compared with the single model, all multi-task models improved the overall F1 scores.
However, Multi-ALL+${\tt DEP}$ compared unfavorably with Multi-ALL even though it was the best PASA architecture.
Therefore, this implies that the dependency type feature between the predicate and its argument is not effective in ENASA.

\subsection{Analysis}


In Figure \ref{analysis}, we compare the PASA results from test sets for each model. 
In Examples (a), (b) and (d), the single model failed to predict correct arguments but the Multi-RNN model correctly predicted arguments.
In Example (a), the single model incorrectly predicted that arguments do not exist in this sentence.
Comparing the training set of each task, although the number of event-nouns is approximately one-third of the number of predicates, the number of {\it kessei 結成} ``organize (event-nouns)" is approximately twice the number of {\it kessei 結成} ``organize (predicates)."
Accordingly, we showed that the Multi-RNN model effectively leverages the information of event-nouns using multi-task learning.

In Example (b), the single model incorrectly predicted that the NOM argument does not exist, but the multi-RNN predicted the correct arguments.
Comparing the training set, there is {\it sayuu 左右} ``determine (predicate)" but not {\it sayuu 左右} ``determine (event-noun)."
However, there are some {\it kagi カギ} ``key (arguments of predicates)" in the PASA training set, and there is one {\it kagi カギ} ``key (argument of event-noun)" in the ENASA training set (Example (c)).
Moreover, in Example (c), {\it dakai 打開} ``break (event-noun)" depends on
{\it kagi カギ} ``key" like {\it sayuu 左右} ``determine (predicate)" in
Example (b); however, no predicate depends on {\it kagi カギ} ``key" in the training set.   
Accordingly, the Multi-RNN model also leverages the arguments of event-nouns and the positional relations between event-nouns and their arguments.

Example (d) is an interesting case in which a predicate {\it kaihi 回避} ``avoid" and its argument {\it sekinin 責任} ``responsibility" are located in the same bunsetsu.
Although this argument type (Bunsetsu) is excluded from the evaluation target in PASA, it is common as a compound noun in ENASA. 
Therefore, the single model wrongly predicted that the ACC argument does not exist, but multi-RNN was able to predict the answer using the specific knowledge of event-nouns. 

In contrast, in Example (e), the single model correctly predicted the answer, but the multi-RNN model failed to predict the correct arguments. 
Multi-RNN incorrectly predicted that the DAT argument does not exist in this sentence.
However, {\it ni に}, a postpositional particle located after an argument, often indicates a dative case. Nevertheless, multi-RNN often predicted a wrong DAT argument by ignoring {\it ni に}.  
Therefore, for DAT analysis, the information of event-nouns adversely affects PASA.

In Example (f), the Multi-ALL+${\tt DEP}$ model correctly predicted the answer, but the Multi-ALL model failed. 
Specifically, Multi-ALL+${\tt DEP}$ correctly predicted that the ACC argument is {\it yakuwari 役割} ``role," which is dependent on {\it hatasu 果たす} ``play." However, the Multi-ALL incorrectly predicted that the ACC argument is {\it kaimei 解明} ``solution."
Similarly, Multi-ALL without syntactic information made many mistakes, including attributive modification, such as Figure \ref{e3}.
Table \ref{rentai} shows the results of the two PASA models for attributive modification instances. 
Multi-ALL+${\tt DEP}$ considerably outperformed Multi-ALL for all cases using dependency features.
Therefore, these results suggest that the dependency type feature is effective for PASA with respect to attributive modifications.

\begin{table}[t]
\centering
\small
\begin{tabular}{l || c c c c} \hline
                      & ALL   & NOM   & ACC   & DAT \\
\hline
Multi-ALL             & 80.31 & 83.37 & 72.16 & 19.48 \\
Multi-ALL+${\tt DEP}$ & 81.83 & 84.67 & 74.41 & 28.31 \\
\hline
\end{tabular}
\caption{F1 scores on the PASA test set with respect to attributive modifications.}
\label{rentai}
\end{table}

\section{Conclusion}
We design a multi-task learning model for predicate and event-noun argument structure analysis.
The experiment results show that the multi-task models outperform the single-task model on the NAIST Test Corpus for both tasks.
Moreover, our model achieves a state-of-the-art result for PASA.
In addition, this is the first work to employ neural networks for ENASA.
In future work, we plan to consider multiple predicates and event-nouns.



\bibliography{naaclhlt2019}
\bibliographystyle{acl_natbib}

\end{document}